# A Framework to Illustrate Kinematic Behavior of Mechanisms by Haptic Feedback

Zhang Q.[1], Chablat D.[1], Bennis F.[1], Zhang W.[2]

Institut de Recherche en Communications et Cybernétique de Nantes
1, rue de la Noë, 44321 Nantes, France
E-mail: {Damien.Chablat, Fouad.Bennis}@irccyn.ec-nantes.fr,
Qinqin.Zhang@eleves.ec-nantes.fr, zhangwei@tsinghua.edu.cn
Department of Industrial Engineering,
Tsinghua University, Beijing 100084, China

**Abstract:** The kinematic properties of mechanisms are well known by the researchers and teachers. The theory based on the study of Jacobian matrices allows us to explain, for example, the singular configuration. However, in many cases, the physical sense of such properties is difficult to explain to students. The aim of this article is to use haptic feedback to render to the user the signification of different kinematic indices. The framework uses a Phantom Omni and a serial and parallel mechanism with two degrees of freedom. The end-effector of both mechanisms can be moved either by classical mouse, or Phantom Omni with or without feedback.

## 1- Introduction

The main purpose of this paper is to introduce a framework to illustrate some kinematic properties by using virtual reality. Virtual Reality is a way for humans to visualize, manipulate and interact with computers and extremely complex data **[1]**. The visualization part refers to the computer generating visual, auditory, haptic or other sensual outputs to the user of a world within the computer.

Commonly, the user gets most of the information that he needs from the visual way, which is also the most convenient way. But under some special conditions, for example, in the robotic simulation area, it will be very useful to add the haptic way to inform the user about the working condition of the robot and what is more important, to warn the user of some dangerous working positions.

The main function of haptic is to make the user feel something that does not exist in reality. One can touch or even manipulate virtual objects. This characteristic makes it a very useful approach in robotic simulation to create an environment closer to reality.

With the intention of illustrating, for the user, kinematic properties of simple robots and by using virtual reality, one application is specially designed with two simple 2-dof linkage manipulators. They are completely haptic manipulators and the user can have the feeling of catching their end-effectors and dragging them. Several different sensual output ways (visual or haptic) have been added in this application to help the user finish the operations more easily, which is what he wants to do.

## 2- Mechanisms under study

In the context of the lecture dealing with parallel mechanisms, we want to introduce basic concept related to their behaviors.

### 2.1- A two-dof serial mechanism

The manipulator under study is a two-bar, revolute(R)-coupled linkage as displayed (Figure **1**). The position of its end-effector $P$ is entirely defined by the lengths $L_1$, $L_2$ and the joint variables $\theta_1$ and $\theta_2$.

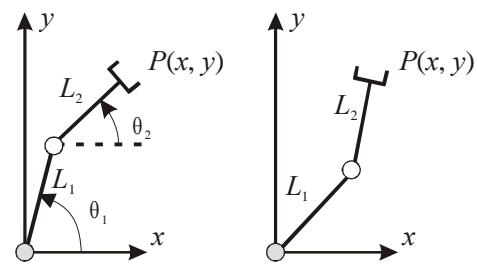

**Figure 1:** A two-dof mechanism

This mechanism admits only one solution to the direct kinematics and two solutions to the inverse kinematics. The singular configurations define the boundaries of the Cartesian workspace and can be characterized by $q_2 - q_1 = 0 + k\pi$. The determinant of the Jacobian matrix is equal to

$L_1 L_2 \det(\theta_1 - \theta_2)$. The kinematic behaviors that we want to illustrate in this framework are:
- The boundaries of the Cartesian workspace,
- The two inverse kinematic solutions,
- The non-homogeneity of the Cartesian workspace,
- The notion of aspect to define continuous trajectories with or without obstacles [2],
- The influence of the joint limits (Fig. 2).

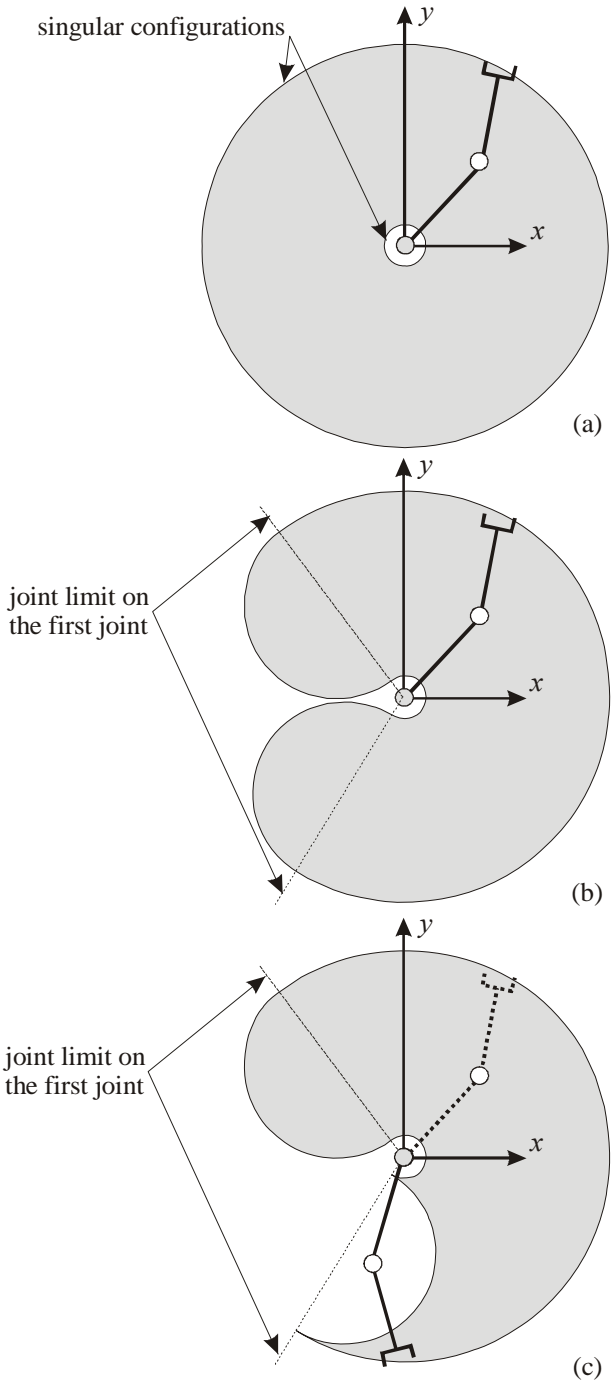

(a)

(b)

(c)

**Figure 2:** Cartesian workspace (a) without joint limits, (b) with joint limits on the first revolute joint and (c) an aspect

The user can move freely the end-effector of the mechanism and change its posture, i.e. change the inverse kinematics solution. When no joint limits are defined, any trajectory can be defined but, in the other case, the Cartesian workspace is not free of singular configuration and, for a given posture, an aspect, i.e. the maximal singularity free region, is smaller than the Cartesian workspace, as is depicted in Fig. 2.

To feel the joint limits, an index is created, whose values is normally equal to zero except in the vicinity of the limits as shown in Fig. 3.

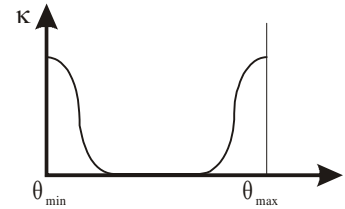

**Figure 3:** Index related to the joint limits

2.2- A two-dof closed-chain mechanism

The manipulator under study is a five-bar, revolute(R)-coupled linkage as displayed (Figure 4). The actuated joint variables are $\theta_1$ and $\theta_2$, while the Cartesian variables are the (x, y) coordinates of the revolute center P. Lengths $L_0$, $L_1$, $L_2$, $L_3$, and $L_4$ define the geometry of this manipulator entirely.

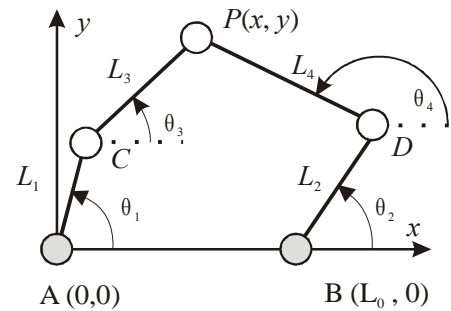

**Figure 4:** A two-dof closed-chain mechanism

The kinematic behaviors that we want to illustrate in this framework are:
- The boundaries of the Cartesian workspace;
- The four inverse kinematic solutions of working modes [3] (Fig. 5);
- The two direct kinematic solutions or assembly modes (Fig. 6);
- The non-homogeneity of the behavior throughout the Cartesian workspace;
- The singular configurations, which can divide the Cartesian workspace.

Two kinematic matrices are used to characterize the behavior of such mechanism [4], matrix **A** is related to the direct kinematics and matrix **B** is related to the inverse kinematics.

$$\mathbf{A} = \begin{bmatrix} L_3 \cos(\theta_3) & L_4 \cos(\theta_4) \\ L_3 \sin(\theta_3) & L_4 \cos(\theta_4) \end{bmatrix} \text{ and }$$

$$\mathbf{B} = \begin{bmatrix} L_2 L_3 \sin(\theta_3 - \theta_1) & 0 \\ 0 & L_2 L_4 \sin(\theta_4 - \theta_2) \end{bmatrix}$$

The determinant of both matrices is computed, det(**A**) and det(**B**). When we multiply the two indices, we define the maximal singularity free regions of the workspace for a given assembly mode and working modes (Fig. **7**), which are called aspects **[5]**. The index is normalized to vary between 0 and 1.

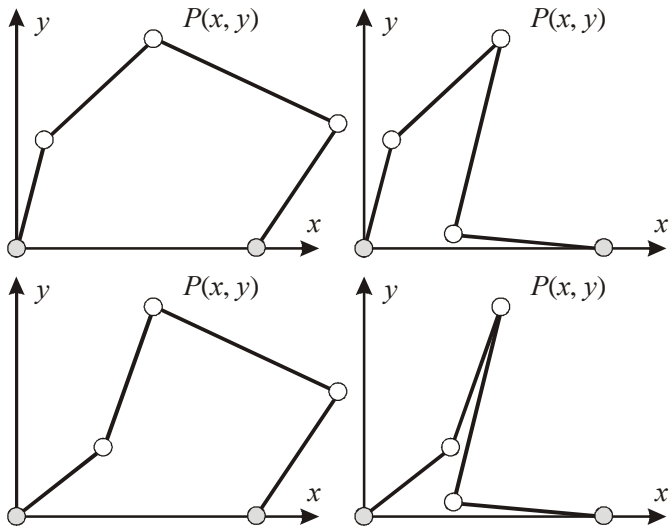

**Figure 5:** The four working modes of the two-dof closed-chain mechanism

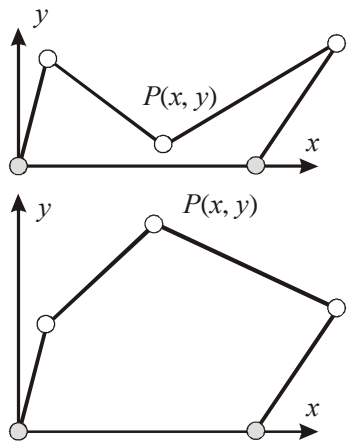

**Figure 6:** The two assembly modes

There are two types of singularities for parallel mechanisms. The first one is associated to the degeneracy of **A** and is located inside the Cartesian workspace. The second one, associated to the degeneracy of **B** defines mainly the boundaries of the Cartesian workspace as shown in Fig. **7**.

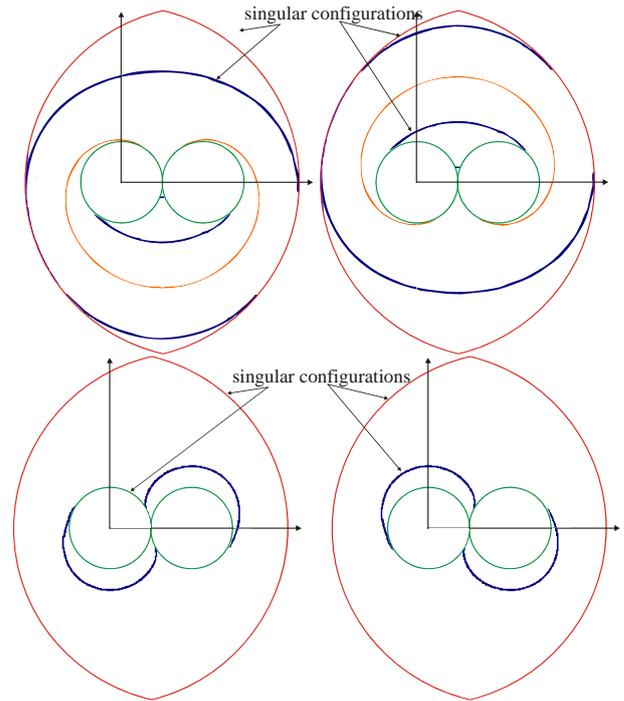

**Figure 7:** Singular configurations for all the working modes

### 3- Basic environment

To realize the incorporation of the sense of touch and control into computer applications through force feedback, it is necessary to have one special input device, called haptic device, and one software environment.

#### 3.1- Hardware: Phantom Omni

The Phantom Omni haptic device (Fig. **8**) enables the users to feel and manipulate virtual 3-dimensional objects. It provides the users with a large workspace with force rendering and it has a 6-dof kinematic capacity. Thus, users can use the stylus of the Phantom Omini device to substitute the mouse and get an experience of touching the virtual objects.

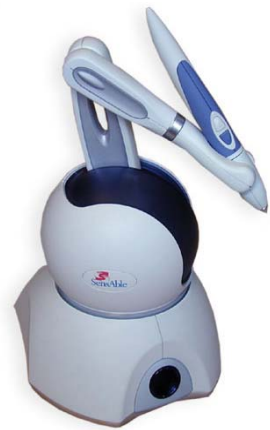

**Figure 8:** The Phantom Omni from Sensable **[6]**

3.2- Software: OpenHaptics[TM] Toolkit

The OpenHaptics[TM] Toolkit gives the users more flexibility to create their own haptic computer applications of special purposes. It has a set of library utilities, which provide both low-level access to the haptic device (Haptic Device API) and high-level haptic rendering (Haptic Library API). The core API is C based but some of the utility libraries use C++. To simplify its graphic programming process, it allows significant reuse of existing OpenGL code and also greatly simplifies the synchronization of the haptics and graphics threads.

3.3- Force rendering

We chose to render by a force feedback the value of the kinetostatic indices. The three main classes of forces that can be simulated are (i) motion dependent, (ii) time dependent, or (iii) a combination of both. We will focus on motion dependent forces, which are suitable in the OpenHaptic library, i.e., spring, damper, coulombic friction and inertia. From this set of haptic feedback, we have chosen the coulombic friction, which simply opposes the direction of motion, with a constant magnitude friction force. The coulombic friction force can be represented by the equation $\mathbf{F} = -c\ \text{sgn}(\mathbf{v})$; where $\mathbf{v}$ is the velocity of the end-effector and $c$ is the friction constant, which can be associated with a kinematic index as the determinant. (Viscous friction + Static and dynamic friction).

4- Application

This application is an example of the incorporation of sense of touch and control through force feedback. It is mainly designed to provide the user with a way to better feel and understand the kinematic properties of 2-dof serial and parallel robots with the help of its graphic rendering and 3-dimensional force rendering in a virtual reality environment.

Besides the basic functions of displaying and adjusting the parameters of the two manipulators, it attempts to give the user different choices to manipulate them and to perceive their working conditions in every corresponding position. For example, if one manipulator is reaching its singular position, with the mouse, the user can only see the change of the color of the manipulator and the trajectory; while with the phantom stylus, the user gets one more input of a friction force feedback which is proportional to the determinant(s) of the manipulator.

4.1- Graphical interface

The graphical interface of this application is composed of two parts: an interactive display section (Fig. **9**) and a set of control panels (Fig. **10**).

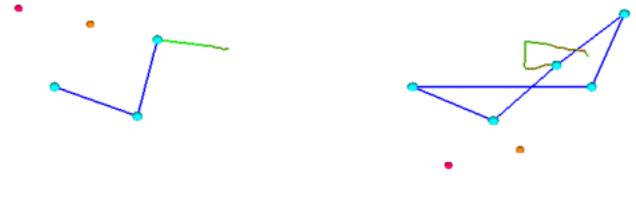

**Figure 9:** The interactive display section of the application

The interactive display section contains one 2-dof serial manipulator on the left and one 2-dof parallel manipulator on the right. There are generally two ways for interaction: the mouse and the phantom stylus.

By using the left button, you can catch and drag the end-effector of the manipulator; by using the middle button, you can zoom in (out) the manipulator; and by using the right button, you can change the view angle and rotate the whole view. But for the stylus, it is only possible to catch and drag the end-effector with its blue button.

There are four additional points in this area, which are the original (yellow) and target (pink) points for the two manipulators. The user can use these four points and the trajectory to evaluate the efficiency of their operations.

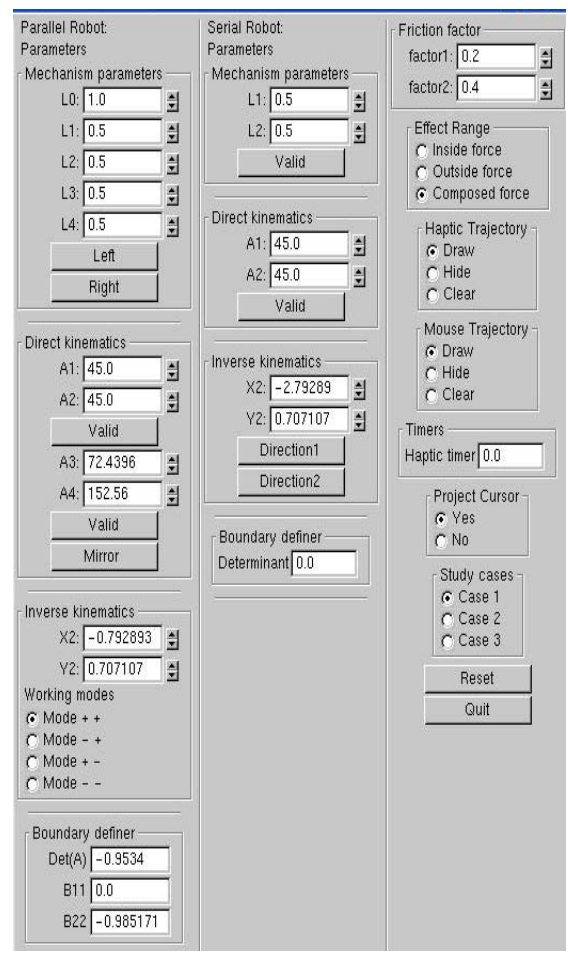

**Figure 10:** The graphic interface of the application

In the control panel section, the only way for interaction is the mouse. There are panels displaying the parameters, direct kinematics, inverse kinematics, and boundary definers of the two manipulators. In addition, there are also panels defining the friction factors, the working and assembly modes, and there are panels enabling one to chose between a hidden, a drawn or a cleared trajectory. The working modes of the manipulators can be chosen and all the angular values can be modified to change the postures and dimensions of the manipulators.

### 4.2- Different ways of outputs

To provide the user with more comprehensive ways to understand kinematic properties, this application makes use of multiple sensory outputs: color, dimension (which are visual approaches) and friction forces (which is a haptic approach). Every output is consistent with the others because they are all for the same purpose to indicate the user to be careful of the manipulators' joint limit positions and workspace limit positions. The first and the second sensory output can be produced in using a classical mouse. For the last one, a Phantom Omni or Phantom Desktop is required.

### 4.2.1- Color

The color of the haptic cursor and the end-effector's trajectory will change, as it is in movement, proportionally to the corresponding manipulator's boundary definer(s). The color for the dangerous areas is defined to be red while the one for the safe areas is defined to be green as people's habits. Before manipulating, the user could get a general position evaluation about different areas just by moving the haptic cursor with phantom stylus through all the areas to see the change of color of the cursor. As soon as one starts to drag the end-effector of the manipulator, the color of the trajectory is a good indicator of the situation.

### 4.2.2- Dimension

In this application, dimension is another way to perceive working conditions. In dangerous areas, the size of the haptic cursor will be larger than its size in safe areas so that the user could get warned. The maximum cursor diameter is two times as large as the minimum cursor diameter. So if combined with the changing of color, a big red cursor indicates very dangerous conditions, while a small green one means that at the very position the manipulator can work safely.

### 4.2.3- Friction force feedback

With the help of the touch-enabled product of SensAble Technologies, haptic is introduced to feel and understand the working situation of the end-effector besides the traditional virtual ways. When it reaches a singular position or a limit position, a friction force is sent to the phantom stylus to warn the user about this situation. This feature makes the perception of the manipulator's working boundaries and kinematic properties easier and more concrete because the movement of the manipulator will be blocked in the dangerous areas by the feedback friction force.

The value of the friction force is a composite function of the corresponding manipulator's boundary definer(s):

$$F(d) = \begin{cases} 0 & (f_1 < d \leq 1) \\ \dfrac{c}{f_1 - f_2}(f_1 - d) & (f_2 < d \leq f_1) \\ c & (0 \leq d \leq f_2) \end{cases}$$

In this function, $F$ is the friction force, $d$ is the boundary definer and $f_1, f_2$ are two friction factors.

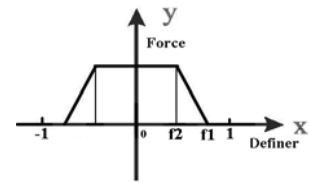

**Figure 11:** The friction force function depending on the boundary definer

There are also three friction force-rendering modes. In the inside force rendering mode, force will generate only when the manipulator is passing its singular position; in the outside force rendering mode, force is only produced when the manipulator is close to its outside working boundary; in the composed force rendering mode, both conditions will cause friction force feedback.

### 4.3- Studying cases

By default, there are three studying cases in the application for the user to choose from. Each studying case defines the positions of one original point and one target point for each manipulator. The task for each studying case is to move the end-effector of the manipulator from the original point to the target point. How to complete this task depends on the positions of the two points and the dimension of the manipulator. In the simpler case, the end-effector can move from the original point to the target point smoothly; or sometimes, it will need to pass some singular position or reach some limit position; while in the most complex case, the task can only be completed after changing the working mode of the manipulator. These studying cases offer the chance to understand more directly different working postures and the uniform base for evaluating the efficiencies of different output channels.

### 4.4- Evaluating functions and examples

The design of this application is to introduce haptic to help the user better understand simple kinematic properties. So it is necessary to add some evaluating functions to evaluate if haptic has positive effect on the perception of working conditions or whether it can help the user in actual manipulations.

In the first step, time is the only realized evaluating function in this application. After selecting the study case and the manipulator, the time it takes between catching the original point and the target point by the Omni stylus and by the mouse can be recorded separately and then the two recorded times can be compared to see if haptic has some positive effect. Along the trajectory, we can evaluate the maximal and minimal values of the kinetostatic index.

The panel associated with the two mechanisms (depicted in Fig. **10**) allows the user to choose the best working mode or assembly mode. However, if the current position of the Omni stylus is not on the right aspect, the user cannot make the trajectory (Fig. **13**), but if we change it, the trajectory can become made (Figs. **12** and **13**).

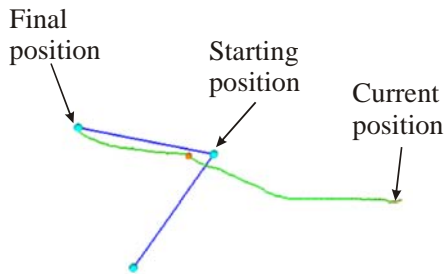

**Figure 12:** Trajectory between two postures for the serial mechanism

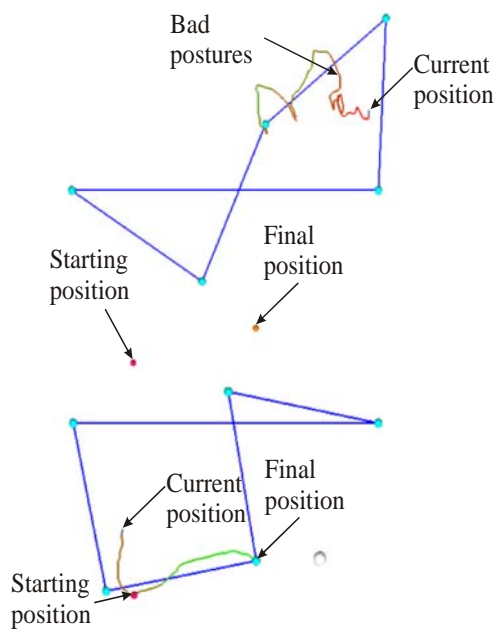

**Figure 13:** Trajectories between two postures for the parallel mechanism

### 5- Conclusions

We have introduced in this paper a new framework for students to feel and understand the properties of mechanisms. Such tools can be used to illustrate some theoretical notions, which are often difficult to explain without robot prototypes. In the context of lecture dealing with virtual reality, this framework allows to us to introduce several types of stimuli (color, shape or haptic feedback). The evaluation of the trajectory was made by the analysis of the duration as well as the minimal and maximal values of the kinetostatic index used.

### 6- Acknowledgments

### 7- References